# Population synthesis for urban resident modeling using deep generative models

Martin Johnsen*[a], Oliver Brandt*[a], Sergio Garrido[a] and Francisco C. Pereira**[a]

[a]*Department of Technology, Management and Economics, Technical University of Denmark, DTU, 2800 Kgs. Lyngby, Denmark*

ABSTRACT

The impacts of new real estate developments are strongly associated to its population distribution (types and compositions of households, incomes, social demographics) conditioned on aspects such as dwelling typology, price, location, and floor level. This paper presents a Machine Learning based method to model the population distribution of upcoming developments of new buildings within larger neighborhood/condo settings.
We use a real data set from Ecopark Township, a real estate development project in Hanoi, Vietnam, where we study two machine learning algorithms from the deep generative models literature to create a population of synthetic agents: Conditional Variational Auto-Encoder (CVAE) and Conditional Generative Adversarial Networks (CGAN). A large experimental study was performed, showing that the CVAE outperforms both the empirical distribution, a non-trivial baseline model, and the CGAN in estimating the population distribution of new real estate development projects.

## 1. Introduction

A crucial component of decision making for sustainable development is to forecast impacts of long-lasting interventions, such as policies, infrastructure investment, and new community areas. Significant work exists, particularly using Agent Based Model (ABM) simulators applied to different research areas such as transport, sociology, and ecology [26]. The benefit of ABM approaches is that simulation can reproduce complex interactions and decision making chains for each agent (e.g. location decisions for job→house→school) and groups of agents (e.g. traffic flow).

In the vast majority of cases, ABMs aim towards large spatial contexts (e.g. city, region, country), thus considering coarse representations at the higher resolution level. For example, agents' home locations are represented at most at the zone level, instead of specific building location. This is more than sufficient for policies and investments at the coarser regional/urban level, for example planning for sustainable urban energy development [32], but becomes inadequate for planning large investments at a much smaller scale. From a real estate development and infrastructure planning perspective, understanding the specific populations on a basis of individual building, together with their usage of resources, mobility, and space becomes a necessity. The main purpose of knowing the development of a population and its characteristics, is to serve the people properly with the necessary amenities. Having a good level of service has been proven to improve life quality in, among others, transport studies [35], and economic studies [12].

In this work, we focus on the first and fundamental step of ABMs: population synthesis, which consists of accurately modeling the population distribution for the area of study. Specifically, we focus on the real estate project development level. This means, in practical terms, that we must synthesize agents at a very high level of detail, taking into account characteristics of, for example, property type and the house topology.

Just as with urban models, which consider surrounding areas as exogenous (e.g. other cities, regions, countries), we will consider our study area in isolation, i.e., we will not account for other areas in the same region, and hence ignore overall supply and demand on the real estate market. While that would ultimately be the most robust and precise method, because it accounts for all systemic effects, it is in practice unrealistic, due to limited data in our region of study.

Using real data from Ecopark Township, a real estate development project in Hanoi, Vietnam, we study two machine learning algorithms from the deep generative models literature to create synthetic agents used in ABMs. Particularly, we use a Conditional Variational Auto-Encoder (CVAE) and Conditional Generative Adversarial Networks (CGAN).

## 2. Literature review

For background and contextualization of this paper, we will focus on two major topics: Population Synthesis techniques and Machine Learning models for real estate research and practice.

### 2.1. Population Synthesis

Historically, population synthesis has been practiced using a range of different approaches. In this paper, we will focus on several recent developments in the use of deep generative models for population synthesis applications. Previously, iterative algorithms, such as Iterative Proportional Fitting (IPF), have been a practice standard due to their ease of implementation [30]. However, they are ultimately a heuristic method that simply reproduces the empirical distribution through expansion factors, being overly sensitive to the data distribution of the sample used for synthesis. For example, if some specific combination of properties is absent or underrepresented (e.g. people above or below a certain age, income, etc.), it will strongly bias the outcome. In other words, IPF is not in itself a *model*, given that it cannot generalize beyond the data sample.

---

*These authors contributed equally to this work
**Corresponding author
ORCID(s):

To address this issue, simulation-based approaches have been applied, such as a Gibbs-sampling method proposed in Farooq et al [13], which showed good performance in low dimensional problems (e.g. under 10 population characteristic variables), but having significant computational challenges with higher dimensionality. High dimensionality was partly addressed subsequently by the Bayesian Network from Sun and Erath [34]. However, their work relies on either knowing the topology of the network graph or on finding it through some discovery algorithm which does not scale to high dimensional data or models with latent variables. The use of Hidden Markov Models (HMMs) was also proposed by [31] as another method of synthesizing a population, where each attribute represents a state, and all attributes are sequentially sampled for all individuals.

More recently, deep generative models [18] have proven effective with large-scale generative modeling problems. In a population synthesis setting, deep generative models have mainly been proposed to synthesize transport data by [6] and [8], and medical data by [10] and [39] until now. [7] proposes to use a Variational Auto-Encoder (VAE) to synthesize transport data and successfully generate populations in a high-dimensional setting. Results from [15] show how both a VAE and a Generative Adversarial Network (GAN) with Wasserstein distance can generate zero-samples, i.e., create out-of-sample agents, which makes these types of generative models even more exceptional. [4] trains a GAN which is capable of reconstructing agents described by both tabular and sequential data in a mobility setting. [8] proposes the application of a Conditional Variational Auto-Encoder (CVAE) to estimate the joint distribution of travel preferences conditional on socio-economic and external variables to investigate how transport preferences evolve. Given the importance of the VAE and the GAN algorithms for this paper, we will describe them in more detail in section *Methodology*.

The Variational Auto-Encoder (VAE), introduced by [20], successfully fit and model distributions over large data sets. Building on the development of the VAE, [33] proposes a Conditional Variational Auto-Encoder (CVAE). A CVAE models the distribution of a high-dimensional output space as a generative process, which generates an output, $x$, conditioned on additional input features[1], $c$. Since the initial paper on VAEs, many extensions and modifications have been proposed to optimize performance. As an example, the encoder and decoder models can be modified to be other Neural Network structures like Recurrent Neural Networks (RNNs). [36] gives an overview of the latest developments in autoencoder-based learning and describes three mechanisms to ensure latent representation of features and their distributions. Furthermore, [21] gives an overview of different frameworks and tasks where deep generative models, in particular VAEs, are proved applicable. Two such areas are representation learning and artificial creativity. In conclusion, [21] states that VAEs are one of the few frameworks in current literature that proves to be efficient in learning latent variables and synthesis.

---
[1] In machine learning, the term feature is often used to describe what other literature refer to as a variable or an attribute.

The Generative Adversarial Network (GAN) was introduced by [19], as an alternative way to train generative models. Apart from the GAN, a large number of variations have been proposed. For example, changing the divergence measure minimized by the GAN loss functions from the Jensen Shannon divergence to the Wasserstein distance can improve the stability during training and relieve common problems, such as mode collapse [3]. Later, a conditional version of the GAN was introduced by [25], namely the Conditional Generative Adversarial Network (CGAN). The CGAN is constructed by feeding conditional features to both the generator and the discriminator.

### 2.2. Machine learning in real estate
Machine learning (ML) research within the real estate domain is mainly focused on two applications; predicting housing prices and finding investment opportunities. Other research within the real estate domain is focused on assessing construction speed, mobility patterns, and customer targeting, however, not all using machine learning methods. A few articles were found on occupancy-prediction, however, mostly using aggregate building features or indoor environmental data.

Housing price prediction is the most common application of machine learning within real estate. Models range in complexity from regression models to complex architectures like Convolutional Neural Networks (CNNs) and bi-directional Long Short Term Memory (bi-LSTM) models. [40] uses a CNN to score the aesthetics of pictures from online housing ads and combines this score with basic property attributes to predict the house price using an eXtreme Gradient Boosting (XGBoost) regression model. [5] claims to outperform prior state-of-the-art models by including other similar properties in the neighborhood as model input. They develop a K-nearest similar house sampling (KNSHS) algorithm to find similar nearby properties and inputs sequences of the KNSHS-result and the current to-be-valued property into a bi-LSTM model. The features extracted from the generated sequences are used to predict the house price in a fully connected layer.

The research on investment opportunities includes models to predict which areas of a city are likely to experience gentrification, proposed by [1], and models to predict whether properties are listed below market price, proposed by [17]. The former uses a Random Forrest Classifier while the latter assess several different algorithms such as K-nearest neighbor (KNN), Support Vector Machine (SVM), and a Neural Network (NN).

Creating customer target groups from Social Media profiles by classifying segments within real estate, parenting, and sports, is proposed by [24]. The selected targets help merchants to identify target customers and plan social media strategies. The paper uses a deep Neural Network to classify the different target customers, given features scraped from a particular social media platform.

Occupancy prediction and energy usage prediction is a broadly defined problem which can be solved from many different perspectives. [28] predicts commercial building energy consumption using descriptive features of buildings, such as the size of the building and the number of employees working in the building. [29] develops an indirect-approach-based occupancy prediction model. This model predicts occupancy using machine learning and indoor environmental data with a focus on privacy. Both a decision tree model and a hidden Markov model is proposed. A third approach is presented by [22], who uses a CGAN with Wasserstein distance as loss function to predict demand-side electricity consumption. They do so, by training the CGAN on electricity consumption denoted by 30-minute intervals for small and medium enterprises (SMEs). The generator can generate realistic electricity consumption after training.

### 2.3. Other existing ML applications in real estate

The academic community has not been exhaustively exploring ML applications within real estate, however, companies and startups are using ML to provide services and applications within the industry. Companies provide a range of different applications under the term Property Technology, or PropTech for short. However, to our knowledge, none of the services offered provides a similar concept to urban resident modeling for potential customer acquisition and planning, as proposed in this paper. A brief description of the four most relevant (yet not similar) applications is provided below. For a more comprehensive description, [27], [16], and [11] provide a good overview.

- Compass predicts purchases for agencies: Compass operates a sales listing site focused on major US cities, primarily on the East coast. The company claims to predict when customers are most likely to buy a property based on their search history and notifies the sales agencies whenever a customer is likely to buy[27].

- Sidewalk Labs generate design ideas: The Alphabet company has created a generative design tool that can produce "millions of planning scenarios"[37], given a wide range of foundational information. The tool is meant to help planners facilitate objectives and trade-offs in the best possible manner.

- CityBldr finds the next investment: Finding multi-property development sites can take anything from days to several months. CityBldr uses AI to find suitable real estate sites in seconds and ranks opportunities based on specific parameters[9].

- Localize provides transparency: Localize mainly operates in New York City using AI to provide transparency for property buyers. The company offers detailed knowledge about actual lighting in an apartment, commuting times, parking facilities, etc.[23].

## 3. Methodology

In a population synthesis application, the objective is to enable sampling of a synthetic target population, $\hat{X}$, that resembles a given real and known population, $X$. Specific types of models that can generate data, which resembles real data, are called Generative Models. Deep generative models have proven to obtain high performance in a wide variety of generative tasks, from image [38] to text generation [14]. Generative Adversarial Networks (GANs) and Variational Auto-Encoders (VAEs) are examples of deep generative models that have proven to perform well in creating synthetic agents in population synthesis applications. The two methods fit the full joint distribution for high-dimensional data sets, in contrast to other traditional generative models. Building on prior research, we are introducing the CGAN and CVAE to perform population synthesis on urban resident features to generate synthesized urban residents conditioned on property-specific features.

### 3.1. Generative Models

In contrast to discriminative models, generative models are built to reconstruct the data of interest. A discriminative model solving a standard classification problem can be considered a direct mapping, where an instance of $x$ is used to predict $y$ given $p(y|x, w)$. In generative models, we are trying to approximate a mapping using a significantly different approach. The models are learning an underlying distribution, represented by latent random variables, from which the data originates. This enables generating synthetic data that resembles the real data, potentially being images, text, or urban residents for a real estate project. The intuition follows a famous quote by Richard Feynman, "What I cannot create, I do not understand."[2]

Generative models are characterized by taking samples from a probability distribution (often Gaussian), $z$, and transform them through a generative model, $\theta$. Using Gaussian random variables to generate an approximate distribution, $\hat{p}(x)$, we can compare the models ability to reconstruct the true data distribution, $p(x)$. The loss is the difference between the true distribution, $p(x)$, and the approximate distribution, $\hat{p}(x)$.

Population synthesis for urban resident modeling using deep generative models

The generative properties force the model parameters to represent some underlying structures of the real world, which causes the models to encode hidden or latent patterns implicitly. Both models presented in the next sections are a part of the family of generative models.

### 3.2. Conditional Generative Adversarial Networks

A new way to train a generative model using the concept of adversarial training was introduced by [19]. This approach involves two adversarial models, a generator function $G$ and a discriminator function $D$. These functions are parametrized by neural networks and are trained simultaneously. The generator $G$ captures the distribution of the data, and the discriminator $D$ estimates the probability of the sample being fake (coming from the generator) or real. This way, the generator learns to generate plausible data, and the discriminator learns to distinguish between fake and real samples.

For $G$ to learn the distribution $p_g$ over data, G is first initiated with a sample from a prior noise distribution $p(z)$, then G transforms the sample to a realistic agent. This way $G$ builds a mapping function from the prior noise to data space using a Neural Network.

The adversarial network $D$, called the discriminator is simply a binary classifier. $D$ is supplied with an agent, either one from training data (real) or one generated by $G$ (fake). Formally, $D$ outputs a value between 0 and 1, representing the probability $D(x)$ of $x$ coming from the data. $D$ is trained to maximize this probability of correctly labeling both the agents coming from $G$ and training data. Simultaneously, $G$ is trained to minimize $ln(1 - D(G(z))$. In game theory terminology, $D$ and $G$ are playing a MinMax game, with a value function $V(G, D)$, described by:

$$\min_G \max_D V(D, G) = \mathbb{E}_{x \sim p_{data}(x)}[\ln D(x)] + \mathbb{E}_{z \sim p_z(z)}[\ln(1 - D(G(z)))] \quad (1)$$

where the first term: $\mathbb{E}_{x \sim p_{data}(x)}[\ln D(x)]$ denotes the expected value of the log-probability that $D$ assigns to real data. This term is maximized by $D$. The second term: $\mathbb{E}_{z \sim p_z(z)}[\ln(1 - D(G(z)))]$ expresses the objective of $G$ to minimize the logarithm of one minus the probability of $D$ labeling an agent generated by $G$ to be real. Due to the value function in equation 1, the loss functions for $D$ to minimize the loss $L_D$ can be extracted as:

$$L_D = -[\ln D(x^{(i)}) + \ln(1 - D(G(z^{(i)})))] \quad (2)$$

while the loss function for $G$ is:

$$L_G = \ln(1 - D(G(z^{(i)}))) \quad (3)$$

For every data point $i$, equation 2 and 3 can be maximized simultaneously.

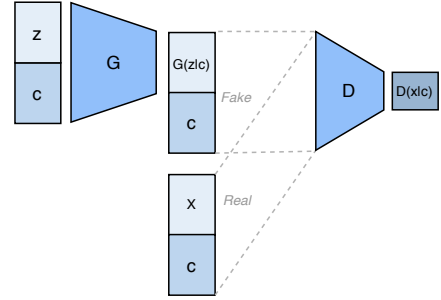

**Figure 1:** High-level architecture of the Conditional Generative Adversarial Network.

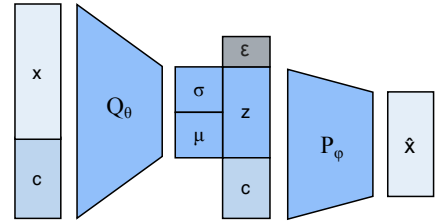

**Figure 2:** High-level architecture of the Conditional Variational Auto-Encoder.

The GAN model can be extended by conditioning on a set of exogenous variables, $c$, as illustrated in figure 1. The conditioning is carried out by feeding $c$ as an additional input to both $D$ and $G$. For $G$ this is done by combining $c$ with the prior input noise $p_z(z)$. With this additional input $c$ the objective function of the MinMax game is now:

$$\min_G \max_D V(D, G) = \mathbb{E}_{x \sim p_{data}(x)}[\ln D(x|c)] + \mathbb{E}_{z \sim p_z(z)}[\ln(1 - D(G(z|c)))] \quad (4)$$

### 3.3. Conditional Variational Auto-Encoder

The architecture of a CVAE is similar to the VAE with an encoder network and a decoder network. However, in the CVAE, the encoder and decoder are conditioned on additional attributes, $c$. The encoder is described by $Q_\theta(z|x, c)$ and the decoder is described by $P_\phi(x|z, c)$. $Q_\theta$ and $P_\phi$ are essentially two composite Neural Networks that mirror each other. $Q$ projects the input $x$ and conditional features $c$, into a latent space, and $P$ reconstructs a synthetic agent, resembling the true agent, given conditional features and the bottleneck layer, $z$. The conditional features are hence fed into the network both at $Q$ and $P$ as figure 2 illustrates.

The bottleneck dimension $D_z$ is oftentimes smaller than the original input dimension $M$ however, alternative versions where $D_z > M$ can also be applied. The lower-dimensional bottleneck should in theory ensure a representation of the true distribution in a lower-dimensional space, ideally close to the true underlying dimension. The conditional variables are included as illustrated in figure 2.

Compared to a VAE, the CVAE is an extension that includes additional conditioning variables, $\boldsymbol{c}$, as an input to the network. It affects the estimation of the joint probability distribution since this now includes the conditional variables, $P(\boldsymbol{x}|\boldsymbol{c})$.

During training, the CVAE is optimized using a loss function, which combines the cross-entropy loss and Kullback-Leibler (KL) divergence. The cross-entropy loss, described by

$$CE(\boldsymbol{x}, \hat{\boldsymbol{x}}) = -\sum_{i=1}^{n} \left[ \boldsymbol{x_i} \log \hat{\boldsymbol{x}_i} + (1 - \boldsymbol{x_i}) \log(1 - \hat{\boldsymbol{x}_i}) \right] \quad (5)$$

measures the reconstruction loss of the decoder, while the KL divergence, described by

$$D_{KL}[Q_\theta(\boldsymbol{z})||P_\phi(\boldsymbol{z})] = -\frac{1}{2} \sum_{k=1}^{D_z} (1 + \log \boldsymbol{\sigma_k} - \boldsymbol{\mu_k}^2 - \boldsymbol{\sigma_k}) \quad (6)$$

measures the divergence between the latent variable distribution and the Gaussian prior, where $Q_\theta(\boldsymbol{z}) = \mathcal{N}(\mu, \sigma)$ and $P_\phi(\boldsymbol{z}) = \mathcal{N}(0, I_{D_z})$. Note that $\mu$ and $\sigma$ are outputs of Q and per definition the approximate posterior.

Equation 5 and 6 lets us define the final loss function as the minimization problem defined in equation 7.

$$\min_{\theta, \phi} \mathcal{L}(\theta, \phi) = CE(\boldsymbol{x}, \hat{\boldsymbol{x}}) + \beta D_{KL}[Q_\theta(\boldsymbol{z})||P_\phi(\boldsymbol{z})] \quad (7)$$

where $\beta$ is a hyperparameter weighting the regularization term, formalized by the KL divergence.

Neural Networks learn through the process of backpropagation. To enable backpropagation through both $Q$ and $P$, the stochasticity has to be separated from the learned parameters. This is a technique known as the reparameterization trick. Technically, the latent variable $\boldsymbol{z}$ is created from $\sigma$, $\mu$, and $\epsilon$, where $\epsilon \sim \mathcal{N}(0, I_{D_z})$, and hence $\boldsymbol{z}$ is not sampled directly, which lets us backpropagate through the full network.

### 3.4. Empirical distribution tables

To compare the generative performance of the models developed in this paper, empirical distribution tables are used as a baseline model. Empirical distribution tables are essentially the observed distribution, given a combination of conditional features. The distributions are all discrete since the data is split into discretized categorical features. Typically, using the empirical distribution on a lower-dimensional data set will provide a strong baseline model because the number of possible value combinations becomes small enough to be fully covered by the original data set.

Empirical distribution tables can be used to assess the generative performance over the marginal, bivariate, and trivariate distributions of the model outputs. All the distribution tables are discrete distribution tables with bins equivalent to the bins defined in the data set. When assessing the performance of empirical distribution tables, it is crucial to keep scalability in mind. On data sets with moderate combinations between conditionals and where test and train data reflect the same distribution, the empirical distribution might perform superior to any model. However, as the number of combinations increases, they become infeasible and poor in performance.

Marginals are sampled by making a distribution table containing all combinations of the conditionals in the training set. Formally, we can describe the marginal probability distribution as $p_X(x)$. Using the computed distribution tables, we can sample a synthetic population based on the empirical distributions by drawing samples from the combinatorial tables. This simple approach is used to generate a baseline population. Since the combinatorial tables are made from the training set (68% of the full data set), it is possible to encounter combinations of conditionals in the test set not seen in the training set. If such unknown combinations are met, the sample is drawn from the overall distribution of the training set.

Bivariate and trivariate probabilities are computed by considering the joint probability distribution over two or three variables, $p_{X,Y}(x, y)$ and $p_{X,Y,Z}(x, y, z)$ respectively.

## 4. Experiments

The goal of population synthesis is to enable the generation of realistic agents characterized by specific attributes within a given domain. In this paper, we have taken an applied approach to population synthesis within the urban housing and real estate domain, which we are describing as urban resident modeling. By using characteristics from real urban residents in the city of Ecopark, denoted by current property owners, we are training generative models to resemble realistic urban residents. However, we do not only wish to enable generation of urban residents in the city of Ecopark. We want to generate urban residents conditioned on specific property-attributes, such as property type (villa, townhouse, apartment), size, price, etc. The purpose of such an application is to enable city planners and sales teams to have qualitative data about residents available for decision support, on a case-by-case basis, given the real estate project considered. With the population synthesis application proposed in this paper, it is possible to generate residents even before a project is built and occupied, given the conditional aspect of the models.

Ecopark township focuses on smart city development and is currently hosting a growing and diverse population of approximately 21,000 residents from almost 50 different countries. Figure 3 provides a visual representation of Ecopark Township of which one third is currently developed.

The experiments are organized as visualized in figure 4. First, we train a model using data consisting of both conditional features and output features (the reader is referred to table

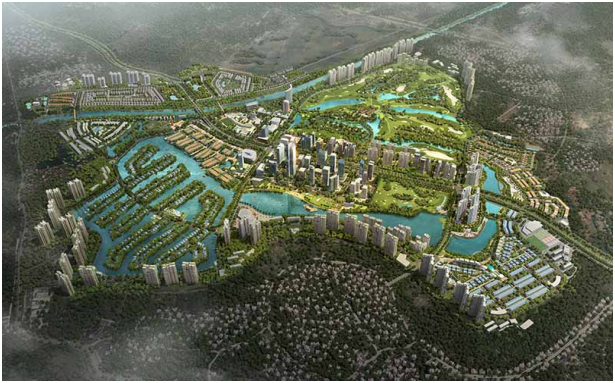

**Figure 3:** Ecopark township is located approximately 15 km outside of Hanoi, Vietnam. The total area is 500 Hectares.

| Feature # | Feature | Type | Categories | Description |
|---|---|---|---|---|
| 1 | Age | Output | 8 | Age of the customer |
| 2 | Gender | Output | 2 | Male or female customer |
| 3 | Nationality | Output | 12 | Nationality of the customer |
| 4 | Investor | Output | 2 | If the customer owns more than 1 property |
| 5 | Prior_home | Output | 12 | District where the customer used to live |
| 6 | Distance_phase1 | Conditional | 4 | Distance to city center 1 in Ecopark |
| 7 | Distance_phase2 | Conditional | 4 | Distance to city center 2 in Ecopark |
| 8 | Distance_greenfield | Conditional | 4 | Distance to Greenfield school in Ecopark |
| 9 | Sales_price | Conditional | 11 | Sales price of property |
| 10 | Size | Conditional | 5 | Size of property |
| 11 | Floor | Conditional | 5 | Apartment floor (only for apartments) |
| 12 | Property_type | Conditional | 3 | Type of property; villa, townhouse, apartment |

**Table 1**
Overview of features. Type indicates whether the feature is a part of the output features or conditional features. During model sampling, output features are generated for each synthesized urban resident, and conditional features are used as an input to influence and condition the characteristics of the urban residents. Note that all features, both output and conditional features, are discretized to categorical features. In total, 6,893 observations are obtained.

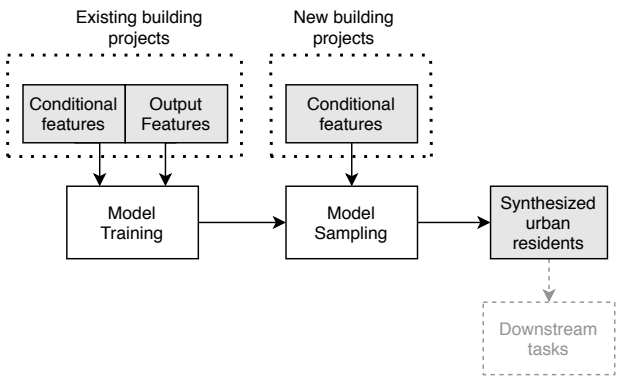

**Figure 4:** Visualization of experiment setup, where white and grey boxes represent models and data respectively.

1 for a description of conditional and output features). Secondly, we can sample synthesized urban residents from the learned distribution using conditional features for a new real estate project. Lastly, the sampled output features can potentially be used in further downstream tasks.

### 4.1. Data description

To generate urban residents in Ecopark, the CVAE and CGAN need a data set consisting of two feature-pairs; features to be generated and conditional features. Throughout the rest of the paper, we will refer to the first type of features as *output features* and the second type of features as *conditional features*. For an illustration of how the CGAN and CVAE are using the output features and conditional features during training, the reader is referred to figures 1 and 2.

The output features are anonymous demographic information on the current property owners in Ecopark. By training a generative model to generate urban residential property buyers defined by demographic characteristics, Ecopark will receive a detailed qualitative overview of the potential residential buyers, which can be used as support in strategic decisions around new projects. Furthermore, urban residents can be used as input to later downstream models. Downstream models are not implemented in this paper but can be valuable for both sales teams and city planners.

The conditional features are not person-specific but rather property-specific, as the conditional features are supposed to be input for the models upon generation of new urban residents. The conditional features consist of essential property attributes. Intuitively, we would expect future urban residents to have correlations between certain demographic features and specific property attributes, like type, location, etc. An overview of both output features, conditional features, as well as a short description, is provided in Table 1.

All features being used in the models, both output and conditional, are discretized to categorical features. There are two main reasons why categorical features are used throughout the paper. Intuitively, it makes sense to restrict the model in terms of the output. If the variables are not categorical, the model has more freedom to generate unrealistic values. Furthermore, many of the features are already categorical by nature. In total, 6,893 observations are obtained from real estate sales in Ecopark from 2008-2019.

As the model-performance can be heavily dependent on the number of combinations between the variables, we consider two versions of the data set in the next sections. Different thresholds of the categorical bins define the two versions. For example, age intervals and price intervals can be of any arbitrary size. The two versions hold the same features as defined in table 1, but with different categorical dimensions:

1. *Original* data set, which consists of bin sizes as defined in the categories column in table 1. The Original data set gives an output dimension of 36 and a conditional dimension of 36.
2. *Extended* data set increases the number of categories in selected features; age, distances, price, size, and floor. Extending the categories gives an output dimension of 45 and a conditional dimension of 49.

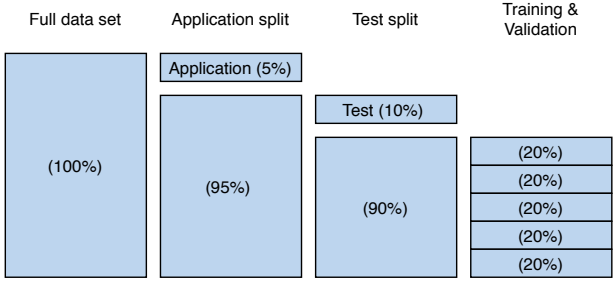

**Figure 5:** Visualization of splitting the data set into application, test, training and validation. We are using cross-validation with $K = 5$ on the training and validation data set, as illustrated in the figure.

### 4.2. Results and discussion

Before training, the data is split into four distinct data sets; training, validation, test, and application as illustrated in figure 5. An application set is not a requirement. It is, however, a way to demonstrate the real-world potential of this population synthesis application. Suppose Ecopark was to use the population synthesis application as a tool in new real estate projects. In that case, Ecopark should be able to input basic property attributes for a new project, and the model should generate output features for each unit in that defined project. For example, Ecopark could define a project consisting of 200 townhouses, and the model would generate urban residential buyers for each of the 200 townhouses. The defined application set consists of 440 residential buyers (approximately 5% of the total data set), from two high-rises called "Rung Co A" and "Rung Co E." Both high-rises are from the same real estate project. The rest of the data is split into approximately 90% training and a 10% test. The defined data splits leave around 15% of the data unseen across the test and application set. During training, the training data is split into 80% training and 20% validation in a cross-validation loop ($K = 5$). The training set is used to fit the internal parameters to the CGAN and CVAE (weights and biases), while the validation set is used to choose the hyperparameters yielding the best performance.

The CVAE is trained in a grid-search-like style, where a range of models have been trained on different hyperparameters. 72 experiments have been conducted with varying hyperparameters across; batch size, hidden layers, hidden units across layers, bottleneck size, learning rate, optimizers, beta values, activation functions, and number of epochs. The best models are characterized by having one hidden layer and a bottleneck dimension between 25 and 30. Details of the best model architecture are provided later. During training, the validation loss and validation distance measure are calculated at every 100th iteration.

The CGAN is trained in the same kind of grid-search-like style as the CVAE. Likewise, a variety of models have been trained with varying hyperparameters. Varying hyperparameters have been tested across: batch size, hidden layers, hidden units, learning rate, activation function, and the number of epochs. We experiment with 1, 2, or 3 hidden layers in both the generator and the discriminator. The validation distance measure is calculated at every 100th iteration to observe the SRMSE during training.

To compare and assess the models, we calculate the Standardized Root Mean Squared Error (SRMSE) between the empirical distribution and the model output. Considering both marginal, bivariate, and trivariate distributions, it becomes apparent which model approximates the marginal and partial join distributions over the data set best. The SRMSE is described in[8]:

$$SRMSE = \frac{RMSE}{\bar{\pi}} = \sqrt{\sum_i \cdots \sum_j (\hat{\pi}_{i..j} - \pi_{i..j})^2 N_c} \quad (8)$$

where $\pi_{i..j}$ are the partial join distributions of variable $i..j$ and $N_c$ is the total number of bins compared. Two other standards metrics are also used to assess the performance; R-squared ($R^2$) and the Pearson correlation coefficient. In the following, the best CVAE and CGAN from all conducted experiments and their performance across the Original and Extended data set are presented and compared to the baseline model. Performance is shown for both validation, test, and application sets, however, the validation performance decides which model architecture is optimal for the CVAE and CGAN respectively. In general, the lower SRMSE-score the better.

The numerous experiments for both the CVAE and CGAN are evaluated by the SRMSE-score on the validation set in the best performing K-fold. The SRMSE is calculated over the marginal distributions for the generated and true samples. Furthermore, we assess the mean, $\mu$, and standard deviation, $\sigma$, of the SRMSE performance over all five K-folds, and the models' ability to generate zero-samples. Zero-samples are defined as a models' generative ability to create synthetic urban residents not encountered among the real urban residents.

From all the conducted experiments using the CGAN, we find the lowest SRMSE score on validation data using a batch-size of 64, a learning rate of 0.001, RMSProp optimizer, one hidden layer containing 1,200 hidden units in both the generator and the discriminator. The number of hidden units examined are; 40, 80, 120, 200, 400, 800, 1,200, and 1,400. Increasing the number of hidden units resulted in a lower SRMSE, until the number of hidden units was above 1,200. This model was trained for 51 epochs, as we found a tendency to overfit the validation data when training for more than 51 epochs. The best performing CGAN is presented in table 2, showing the SRMSE over marginals, $\mu$, and $\sigma$ on both the Original and Extended data.

From the 72 different experiments performed on the CVAE, the best architecture is found to be: one hidden layer with 50 hidden units, a bottleneck dimension of 25, batch-size 32, learning rate of 0.001, RMSProp optimizer, a beta value of 0.5, ELU activation function, and 500 training epochs. More complex architectures result in decreased SRMSE performance. More training epochs result in more zero-samples, but worse SRMSE performance in later epochs, resulting in early stopping. The CVAE does, however, not reveal signs of overfitting, even with many training epochs. The performance on both the Original and Extended data set is shown in table 2.

| Model | Marg. | $\mu$ | $\sigma$ | zero-samples (pct.) |
|---|---|---|---|---|
| | Original data | | | |
| CVAE | 0.463 | 0.564 | 0.084 | **1.020** |
| CGAN | 0.771 | 0.734 | 0.031 | 0.939 |
| Baseline | **0.373** | - | - | - |
| | Extended data | | | |
| CVAE | 0.660 | 0.625 | 0.062 | 2.370 |
| CGAN | 0.850 | 0.802 | 0.047 | **2.550** |
| Baseline | **0.491** | - | - | - |

**Table 2**
Performance of the best CVAE and CGAN on the validation set. Performance is measured in SRMSE over the marginal distributions between generated and true samples. Results are shown for both the Original and Extended data set. $\mu$ and $\sigma$ is shown to provide evidence of stable model training. The best performance in each column is marked in bold.

| | Test | | | | Application | | | |
|---|---|---|---|---|---|---|---|---|
| Model | Marg. | Bivar. | Trivar. 1 | Trivar. 2 | Marg. | Bivar. | Trivar. 1 | Trivar. 2 |
| | Original | | | | Original | | | |
| CVAE | **0.609** | **0.326** | **0.422** | **0.361** | **0.691** | **0.187** | 0.414 | 0.342 |
| CGAN | 0.855 | 0.434 | 0.595 | 0.475 | 0.810 | 0.208 | 0.592 | 0.442 |
| Baseline | 0.663 | 0.383 | 0.498 | 0.444 | 0.712 | 0.275 | **0.352** | **0.271** |
| | Extended | | | | Extended | | | |
| CVAE | 0.796 | 0.412 | **0.462** | 0.418 | 0.861 | 0.456 | **0.517** | 0.421 |
| CGAN | 1.102 | 0.527 | 0.801 | 0.662 | 0.955 | 0.600 | 0.770 | 0.618 |
| Baseline | **0.746** | **0.347** | 0.477 | 0.443 | 0.868 | **0.404** | 0.546 | 0.451 |

**Table 3**
Performance of the best CVAE and best CGAN when sampling on the test and application data set. The models are compared to each other and the baseline using the SRMSE distance over the marginal, bivariate, and trivariate distributions. The bivariate distribution is between age and nationality. Trivariate distribution 1 is between age, nationality, and prior home district. Trivariate distribution 2 is between age, prior home district, and investor. The best performance in each column is marked in bold.

An examination of table 2 reveals that the baseline is beating the generative models, CVAE and CGAN, on the marginal SRMSE over the validation set in the best K-fold. This is true for both the Original and Extended data set, however, several reasons for this behavior can be found in the inherent size of the data set. In high-dimensional problems, the number of combinations between features can be in the millions. A combination is defined as the composition of an agent, given the available feature categories. Table 1 represents the features and the number of categories in the Original data set. Multiplying the number of categories in the output variables gives a total of 4,608 theoretical combinations. In the Extended data set, there are 7,488 theoretical combinations. Given this relatively small number of combinations, the empirical distribution (and thus the baseline) is expected to do very well. As the amount of features increases, the amount of possible combinations increases as well. With a significant number of combinations, the empirical distributions will be practically infeasible. However, for smaller data sets, the empirical distributions will be accurate. With this in mind, the models show a good performance relative to the baseline, as we can consider the baseline to be a non-trivial model in terms of performance. There are two primary reasons for why the generative models are preferable over the baseline, given the results in table 2; first, the generative models are scalable, meaning that Ecopark can add more features while they continue to digitize different services. More features would potentially increase the application-value without decreasing the performance, as would eventually happen with the baseline. Second, both of the generative models can generate zero-samples, which is impossible for the baseline.

Another critical observation is that the models are significantly better at generating the desired zero-samples on the Extended data set, while not losing SRMSE, compared to the baseline across the two data sets. Higher zero-samples without losing too much performance favor the Extended data set over the Original data set in further discussions of model performance and results.

To further analyze model performance, we are looking at the generative performance over the test set and application set, which have not been fitted during training. The performance is extended to include partial joint distributions, namely the bivariate and trivariate distributions in addition to the marginals. Note that testing model performance on the application set demonstrates how Ecopark could use the model in a real-world setting, as we are practically synthesizing future urban residents for a fully defined unseen real estate project. Table 3 provides an overview of the performance over test and application set, across both the Original and Extended data set.

From table 3, it is apparent that the CVAE is on par with the baseline, and even outperforms it on the Extended data on the two trivariate SRMSE metrics. On both the validation, test, and application data set (table 2 and 3), the CVAE is superior to the CGAN. Exactly what is causing this sub-par performance of the CGAN is not immediately clear, however, GANs are in general known to be difficult to train. Training a GAN is akin to a blind search, as there is no perfect indicator of when performance is converging during training. Furthermore, both the GAN and the CGAN are known to suffer from the problem of mode-collapse. Mode-collapse is best described as the CGAN learning a too simplified distribution, where the generator maps multiple inputs to a single output accepted by the discriminator. This causes the CGAN to fail in learning the real distribution, and generate samples with low variety.

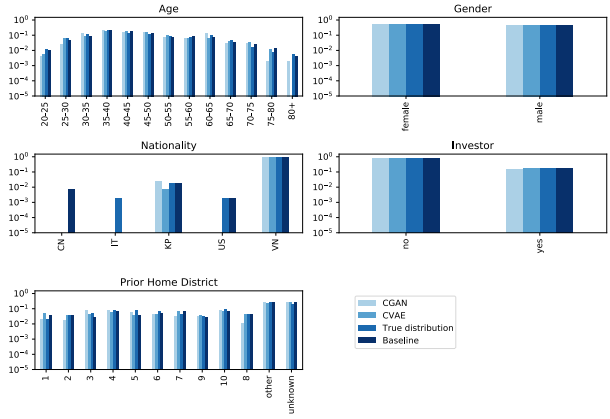

**Figure 6:** Marginal distributions of the five output variables shown on a log scale to emphasize low probability areas. The performance is shown for the test set of the Extended data set.

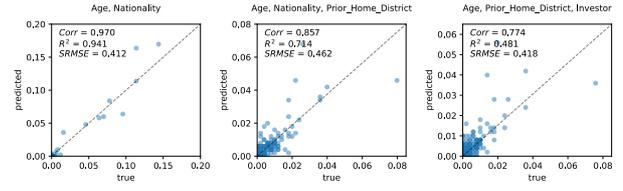

(a) Conditional Variational Auto-Encoder

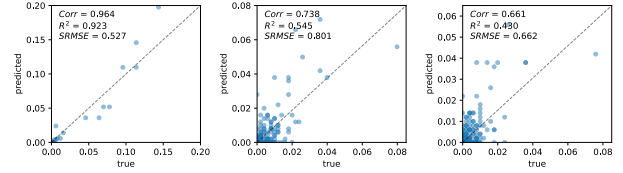

(b) Conditional Generative Adversarial Network

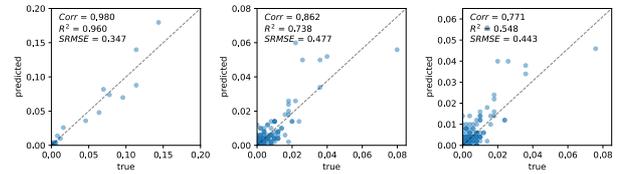

(c) Baseline

**Figure 7:** From left to right; 1) the bivariate distribution between age and nationality, 2) the trivariate distribution between age, nationality, and prior home district, and 3) the trivariate distribution between age, prior home district, and investor. The scatter plot represents the partial joint distribution between the sampled agents from the Extended test set against the real agents from the Extended test set. The axes are denoted in normalized bin frequencies on both the vertical and the horizontal axis.

From table 2 and 3, it is clear that the Extended data indeed results in a higher SRMSE on both the validation, test, and application data. This is expected, as the Extended data set has a significantly higher dimensionality compared to the Original data set. With the increased dimensionality of the Extended data, the CVAE appears to slightly outperform the baseline on the two trivariate distributions on both the test and application data. The fact that SRMSE on the test and application data is more or less on par (except for the bivariate distribution) indicates that the CVAE can generate meaningful residents in a real-world setting.

The marginal distributions are further investigated and visualized in figure 6, showing the samples produced by the CGAN, CVAE, and baseline conditioned on the test set conditionals, and compared to true marginal distribution of the test set. All further plots are shown for the test set of the Extended data set.

In figure 6, all the five output variables are visualized with good approximated marginal distributions across most attributes. Inconsistencies are most evident for the Nationality-attribute, where the low probability values, namely the nationalities US and IT, are not captured by the CGAN or CVAE. It is most likely a general problem with the Nationality-attribute, as many of the nationalities are considerably underrepresented in the data. It is also notable that the baseline samples Chinese (CN) residents without then being present in the test set (true distribution bar). A possible solution to this problem could be to collapse low-frequency nationalities, such as IT and US, into one category, for example, "western countries". Another apparent inconsistency is the CVAE model's inability to generate "80+" year-old residents given the test set conditionals. The inability to generate "80+" characteristics might be a consequence of the same problem as noticed in the low-frequency nationality category. The distributions are plotted on a log-scale, hence the "80+" year segment is, when not plotting on a log-scale, significantly underrepresented in the data. A solution could be to include the low probability age-values in the closest interval, decreasing low-frequency intervals in the tails.

Considering illustrations of the marginal distributions alone is, however, not enough. We also need to assess the partial joint distributions. Figure 7 illustrates the performance of the partial joint distributions to stress-test the generative performance of the models.

In figure 7, the models are assessed on their ability to generate agents across partial joint distributions. Each of the plots represents one model's ability to create agents across two or three variables on the test set. The dots represent a combination between the variables and the frequency of that combination. For example, figure 7 (a), the left-most plot; each dot represents a combination of age and nationality, e.g., 30-35 years and Vietnamese, and the amount of sampled and true agents in each combination. The amount of sampled and true agents in each combination is denoted on the normalized axis. If the amount of sampled and true agents are equal across the two variables compared in the bivariate case, the dots will lie on the dashed diagonal, indicating favorable generative performance. The performance is shown for the Extended test set.

An examination of figures (a), (b), and (c) confirms that the CVAE is superior at capturing the partial joint distributions compared to the CGAN. Both the CVAE and the baseline are scattered systematically around the diagonal, and the CVAE is, as reported in table 3, best across both trivariate distributions. The partial joint distributions are plotted for the application set in appendix *Partial joints in the Extended application data set* Figure 8. Some of the plots show a grid-like structure for the dots, resulting from low-dimensional data with few combinations between variables.

We can generally state that the model performance is satisfying given the various performance metrics; pct. zero-samples and SRMSE across marginal, bivariate, and trivariate distributions. The CVAE is outperforming the baseline in the higher-dimensional Extended data set across the test and application sets. The zero-sampling ability, the scalability, and the generative performance of the CVAE on the Extended data set provide proof for why we conclude the CVAE as the best choice of model for urban resident modeling in a population synthesis application. The best model can be used to generate synthesized urban residents over the application conditionals and demonstrate further downstream tasks. Downstream tasks can be built around the following concepts; mobility-prediction of urban residents, i.e., public transportation patterns and vehicle ownership, occupancy-prediction, i.e. predictive models assessing how many urban residential buyers will move into the property, and clustering models of the urban residents to identify relevant residential groups. These models are, however, left out for future work.

## 5. Conclusion and future work

We have found that deep generative models are a powerful tool to model joint distributions and, therefore, create synthetic agents for Agent Based Models. One of the key elements of our proposed approach is its scalability to higher-dimensional data sets. A researcher or practitioner could eventually have samples from a population with hundreds of characteristics and, given a big enough sample size and proper training of the neural networks, generate samples from the population distribution.

The best performing model was the Conditional Variational Auto-Encoder (CVAE) with an architecture consisting of one hidden layer with 50 hidden units and a bottleneck dimension of 25. The results from table 3 confirm that the CVAE is good at generalizing to test and application data, by beating the baseline model which represents the empirical distribution. The baseline is superior in its generative performance on the marginal distributions on the validation set. However, when considering the performance on test and application set, the baseline is outperformed by the CVAE on the partial joint distributions, which demonstrates the superiority of deep generative models, and in particular the CVAE. On the Extended application set, the CVAE outperforms the baseline on three out of four SRMSE distance metrics and particularly on partial joint distributions. The results also provide evidence that deep generative models are improved when discretizing the inputted variables, in contrast to the baseline. Improved performance on higher-dimensional data indicates that the models are scalable when increasing the data dimensions, which is a highly regarded property in a variety of problems, including urban resident modeling described in this paper.

In further work, we would like to explore how do these agents perform when used in ABMs. Preliminary results of models in transport demand and energy modeling show that they can effectively be used in such tasks. Other interesting areas of possible future work are related to efficient and stable training of the models here presented. The goal of having stable training is to make this process less of a "black-box" for practitioners, who do not know all the specifics of neural network modeling. This future direction can be leveraged by continuous feedback with the machine learning literature where these problems are of vital importance.

# A. Appendix

## A.1. Partial joints in the Extended application data set

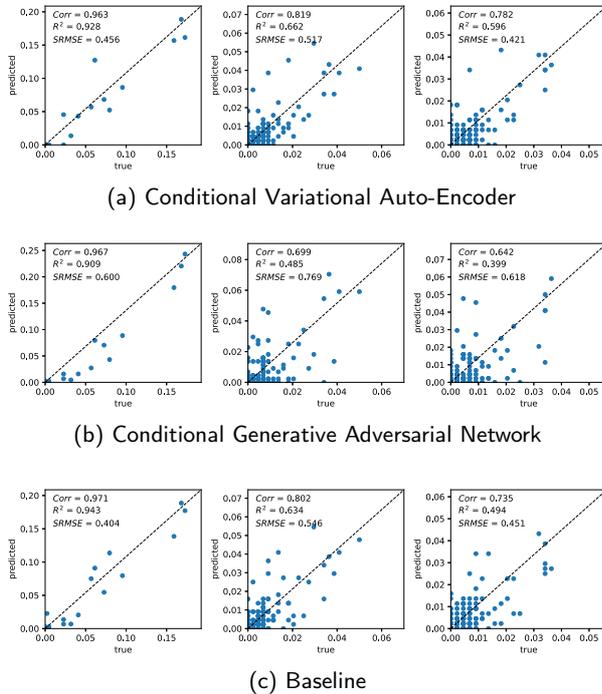

(a) Conditional Variational Auto-Encoder

(b) Conditional Generative Adversarial Network

(c) Baseline

**Figure 8:** Performance of the partial joints in the Extended application data set. From left to right; 1) the bivariate distribution between age and nationality, 2) the trivariate distribution between age, nationality, and prior home district, and 3) the trivariate distribution between age, prior home district, and investor. The scatter plot represents the partial joint distribution between the sampled agents from the Extended application set against the real agents from the Extended application set. The axes are denoted in normalized bin frequencies on both the vertical and the horizontal axis.

# References

[1] Alejandro, Y., Palafox, L., 2019. Gentrification prediction using machine learning. Advances in Soft Computing, Springer International Publishing 978-3-030-33749-0.

[2] Andrej Karpathy, Pieter Abbeel, Greg Brockman, etc.,. Generative models. URL: https://openai.com/blog/generative-models/.

[3] Arjovsky, M., Chintala, S., Bottou, L., 2017. Wasserstein gan. arXiv:1701.07875.

[4] Badu-Marfo, G., Farooq, B., Paterson, Z., 2020. Composite travel generative adversarial networks for tabular and sequential population synthesis .

[5] Bin, J., Gardiner, B., Li, E., Liu, Z., 2019. Peer-dependence valuation model for real estate appraisal. Data-Enabled Discovery and Applications 3. doi:10.1007/s41688-018-0027-0.

[6] Borysov, S., Rich, J., Pereira, F., 2019a. How to generate micro-agents? a deep generative modeling approach to population synthesis. Transportation Research. Part C: Emerging Technologies 106, 73–97. doi:10.1016/j.trc.2019.07.006.

[7] Borysov, S., Rich, J., Pereira, F., 2019b. Scalable population synthesis with deep generative modeling. Elsevier .

[8] Borysov, S.S., Rich, J., 2019. Introducing super pseudo panels: Application to transport preference dynamics. arXiv:1903.00516.

[9] Bumgardner, J., 2020. Citybldr website: https://www.citybldr.com/solutions. URL: https://www.citybldr.com/solutions.

[10] Choi, E., Biswal, S., Malin, B., Duke, J., Stewart, W.F., Sun, J., 2017. Generating multi-label discrete patient records using generative adversarial networks. arXiv:1703.06490.

[11] CIO-Applications, 2019. Top 10 proptech companies - 2019: www.proptech.cioapplicationseurope.com. URL: https://proptech.cioapplicationseurope.com/vendors/Top-PropTech-Companies.html.

[12] Deller, S.C., Tsai, T.H., Marcouiller, D.W., English, D.B., 2001. The role of amenities and quality of life in rural economic growth. American Journal of Agricultural Economics 83, 352–365.

[13] Farooq, B., Bierlaire, M., Hurtubia, R., Flötteröd, G., 2013. Simulation based population synthesis. Transportation Research Part B: Methodological 58. doi:10.1016/j.trb.2013.09.012.

[14] Fedus, W., Goodfellow, I., Dai, A.M., 2018. Maskgan: Better text generation via filling in the arXiv:1801.07736.

[15] "Garrido, S., Borysov, S., Pereira, F., Rich, J., 2019. Prediction of rare feature combinations in population synthesis: Application of deep generative modelling. Elsevier .

[16] Go-Weekly, 2020. Go weekly magazine: The 20 most innovative companies in real estate (or proptech). URL: https://medium.com/go-weekly-blog/the-20-most-innovative-companies-in-real-estate-or-proptech-2e0242b80e32.

[17] Gómez, A.B., Moreno, A.J., Iturrarte, R., Bernárdez, Ó., Afonso, C., 2018. Identifying real estate opportunities using machine learning. ArXiv abs/1809.04933.

[18] Goodfellow, I., Bengio, Y., Courville, A., 2016. Deep Learning. MIT Press.

[19] Goodfellow, I.J., Pouget-Abadie, J., Mirza, M., Xu, B., Warde-Farley, D., Ozair, S., Courville, A., Bengio, Y., 2014. Generative adversarial nets, in: Proceedings of the 27th International Conference on Neural Information Processing Systems - Volume 2, MIT Press, Cambridge, MA, USA. p. 2672–2680.

[20] Kingma, D.P., Welling, M., 2014. Auto-encoding variational bayes. CoRR abs/1312.6114.

[21] Kingma, D.P., Welling, M., 2019. An introduction to variational autoencoders. Foundations and Trends in Machine Learning .

[22] Lan, J., Guo, Q., Sun, H., 2018. Demand side data generating based on conditional generative adversarial networks. Energy Procedia 152, 1188 – 1193. URL: http://www.sciencedirect.com/science/article/pii/S187661021830701X, doi:https://doi.org/10.1016/j.egypro.2018.09.157. cleaner Energy for Cleaner Cities.

[23] Localize, 2020. Localize website: https://www.localize.city/. URL: https://www.localize.city/.

[24] Lv, H.x., Yu, G., Tian, X., Wu, G., 2014. Deep learning-based target customer position extraction on social network. International Conference on Management Science and Engineering - Annual Conference Proceedings , 590–595doi:10.1109/ICMSE.2014.6930283.

[25] Mirza, M., Osindero, S., 2014. Conditional generative adversarial nets. arXiv:1411.1784.

[26] O'Donoghue, C., Morrissey, K., Lennon, J., 2014. Spatial microsimulation modelling: a review of applications and methodological choices .

[27] Rice, M., 2019. 21 ai real estate companies to know. URL: https://builtin.com/artificial-intelligence/ai-real-estate.

[28] Robinson, C., Dilkina, B., Hubbs, J., Zhang, W., Guhathakurta, S., Brown, M.A., Pendyala, R.M., 2017. Machine learning approaches for estimating commercial building energy consumption. Applied Energy 208, 889 – 904. URL: http://www.sciencedirect.com/science/article/pii/S0306261917313429, doi:https://doi.org/10.1016/j.apenergy.2017.09.060.

[29] Ryu, S.H., Moon, H.J., 2016. Development of an occupancy prediction model using indoor environmental data based on machine learning techniques. Building and Environment 107, 1 – 9. URL: http://www.sciencedirect.com/science/article/pii/

S0360132316302463, doi:https://doi.org/10.1016/j.buildenv.2016.06.039.

[30] Saadi, I., Eftekhar, H., Teller, J., Cools, M., 2018. Investigating scalability in population synthesis: a comparative approach. Transportation Planning and Technology 41, 1–12. doi:10.1080/03081060.2018.1504182.

[31] Saadi, I., Mustafa, A., Teller, J., Farooq, B., Cools, M., 2016. Hidden markov model-based population synthesis. Transportation Research Part B: Methodological 90, 1–21. doi:10.1016/j.trb.2016.04.007.

[32] Shi, Z., Fonseca, J.A., Schlueter, A., 2017. A review of simulation-based urban form generation and optimization for energy-driven urban design. Building and Environment 121, 119 – 129. URL: http://www.sciencedirect.com/science/article/pii/S0360132317301865, doi:https://doi.org/10.1016/j.buildenv.2017.05.006.

[33] Sohn, K., Lee, H., Yan, X., 2015. Learning structured output representation using deep conditional generative models, in: Cortes, C., Lawrence, N.D., Lee, D.D., Sugiyama, M., Garnett, R. (Eds.), Advances in Neural Information Processing Systems 28. Curran Associates, Inc., pp. 3483–3491. URL: http://papers.nips.cc/paper/5775-learning-structured-output-representation-using-deep-conditional-generative-models.pdf.

[34] Sun, L., Erath, A., 2015. A bayesian network approach for population synthesis. Transportation Research Part C: Emerging Technologies 61, 49–62.

[35] Todd, L., 2014. Transportation and the Quality of Life. Springer Netherlands, Dordrecht. pp. 6729–6733. URL: https://doi.org/10.1007/978-94-007-0753-5_3053, doi:10.1007/978-94-007-0753-5_3053.

[36] Tschannen, M., Bachem, O., Lucic, M., 2018. Recent advances in autoencoder-based representation learning. CoRR .

[37] Whitney, V., Ho, B., . Sidewalk labs blog: A first step toward the future of neighborhood design. URL: https://www.sidewalklabs.com/blog/a-first-step-toward-the-future-of-neighborhood-design/.

[38] Yan, X., Yang, J., Sohn, K., Lee, H., 2016. Attribute2image: Conditional image generation from visual attributes arXiv:1512.00570.

[39] Yoon, J., Jordon, J., van der Schaar, M., 2019. PATE-GAN: Generating synthetic data with differential privacy guarantees, in: International Conference on Learning Representations. URL: https://openreview.net/forum?id=S1zk9iRqF7.

[40] Zhao, Y., Chetty, G., Tran, D., 2019. Deep learning with xgboost for real estate appraisal , 1396–1401doi:10.1109/SSCI44817.2019.9002790.